\documentclass[letterpaper]{article} 
\usepackage{aaai2026}  
\usepackage{times}  
\usepackage{helvet}  
\usepackage{courier}  
\usepackage[hyphens]{url}  
\usepackage{graphicx} 
\usepackage{natbib}  
\usepackage{caption} 
\usepackage{algorithm}
\usepackage{algorithmic}
\usepackage{amsmath}
\usepackage{amssymb}
\usepackage{soul}
\usepackage{booktabs}
\usepackage{caption}
\usepackage{array}
\usepackage{tabularx}

\definecolor{readgreen}{RGB}{0,150,0}

\usepackage{newfloat}
\usepackage{listings}
\DeclareCaptionStyle{ruled}{labelfont=normalfont,labelsep=colon,strut=off} 
\floatstyle{ruled}
\newfloat{listing}{tb}{lst}{}
\floatname{listing}{Listing}
\usepackage{tikz}
\usetikzlibrary{shapes,arrows,positioning,calc,fit,shadows,backgrounds}

\title{From Kinematics to Dynamics: Learning to Refine Hybrid Plans for Physically Feasible Execution}

\author {
    Lidor Erez\textsuperscript{\rm 1},
    Shahaf S. Shperberg\textsuperscript{\rm 2},
    Ayal Taitler\textsuperscript{\rm 1}
}
\affiliations {
    \textsuperscript{\rm 1} Department of Industrial Engineering and Management\\
    \textsuperscript{\rm 2} Stein Faculty of Computer and Information Science\\
    Ben-Gurion University of the Negev \\ Be'er Sheva, Israel \\
    erezlid@post.bgu.ac.il, \{shperbsh,\ ataitler\}@bgu.ac.il
}

\begin{document}
\maketitle
\thispagestyle{empty}
\pagestyle{empty}

\begin{abstract}

In many robotic tasks, agents must traverse a sequence of spatial regions to complete a mission. Such problems are inherently mixed discrete-continuous: a high-level action sequence and a physically feasible continuous trajectory. The resulting trajectory and action sequence must also satisfy problem constraints such as deadlines, time windows, and velocity or acceleration limits. While hybrid temporal planners attempt to address this challenge, they typically model motion using linear (first-order) dynamics, which cannot guarantee that the resulting plan respects the robot's true physical constraints. Consequently, even when the high-level action sequence is fixed, producing a dynamically feasible trajectory becomes a bi-level optimization problem. We address this problem via reinforcement learning in continuous space. We define a Markov Decision Process that explicitly incorporates analytical second-order constraints and use it to refine first-order plans generated by a hybrid planner. Our results show that this approach can reliably recover physical feasibility and effectively bridge the gap between a planner's initial first-order trajectory and the dynamics required for real execution.

\end{abstract}

\section{Introduction}

Hybrid temporal planning aims to generate execution plans for robotic tasks that satisfy both discrete logical requirements and continuous numeric constraints governed by differential relations, together with temporal constraints. Temporal planners such as \cite{coles2012colin}, \cite{fernandez2018scottyactivity}, and \cite{fernandez2015mixed} have demonstrated strong capabilities in handling complex temporal planning problems. Among these, the Scotty planner~\cite{fernandez2018scottyactivity} is a prominent approach that combines search over plan skeletons with trajectory optimization formulated as a mathematical program. In Scotty, the robotic system is modeled as a first-order integrator over the velocity
\begin{equation}
    \dot x(t) = v(t),
\end{equation}
which enables the formulation of the resulting optimization problem as a Second Order Cone Program (SOCP) \cite{lobo1998applications}. Additional hybrid planners include \cite{chen2021optimal}, which encodes mixed discrete-continuous planning for linear hybrid automata as a Mixed-Integer Linear Program (MILP), and \cite{denenberg2019mixed}, which introduces linear over- and under-estimators to enable reasoning about nonlinear fluents within linear-dynamics planners. Despite their differences, these approaches share a common modeling assumption: they rely on first-order abstractions of system dynamics, where control inputs typically correspond to velocities and no continuity constraints on the inputs are enforced. While this assumption enables tractable optimization formulations such as SOCP or MILP, it fails to capture the second-order nonlinear dynamics that govern real robotic systems, including bounded accelerations and drag effects. Consequently, tracking these first-order plans with a low-level closed-loop controller inevitably induces actuator saturation, leading to large tracking errors and potential instability. In recent work, \cite{taitler2019minimum} treat the plans produced by the mentioned planners as sequences of point-to-point motions and introduce Minimum-Time Validation (MTV), which provides a closed-form lower bound on the time required to move between two way-points under second-order dynamics. This bound makes it possible to identify way-point transitions whose assigned durations are physically unattainable, revealing portions of the plan that cannot be executed in reality. Yet, detecting violations is not sufficient: existing hybrid planners cannot revise a plan within their first-order models to enforce second-order feasibility. 

In this work, we bridge this gap by introducing a neuro-symbolic refinement framework that combines hybrid planning, second-order validation, and reinforcement learning. Starting from a first-order-feasible hybrid plan generated by an off-the-shelf hybrid planner, we represent the temporal and spatial structure as a graph and formulate the refinement task as a continuous-state and action Markov Decision Process (CSA-MDP). A Graph Neural Network (GNN) \cite{zhou2020graph} processes this graph and proposes adjustments to the velocity bounds. An SOCP solver computes the corresponding motion parameters, and the updated trajectory is then validated for physical feasibility via MTV. This yields a closed-loop refinement mechanism capable of computing a feasible-guaranteed second-order trajectory without modifying symbolic preconditions, action ordering, or temporal constraints.

The contributions of this work are as follows:

\begin{itemize}\setlength\itemsep{0em}

    \item We formulate plan refinement as a CSA-MDP 
    Process over a graph-based plan representation, enabling structured reasoning over temporal, spatial, and dynamical relationships and constraints.
    
    \item We introduce the first automated refinement framework that transforms first-order feasible plans into physically valid trajectories by integrating closed-form second-order feasibility analysis with reinforcement learning.

    \item We provide an empirical evaluation across multiple hybrid planning domains, demonstrating that first-order plans are physically infeasible under second-order dynamics, and that our refinement approach consistently achieves second-order feasibility.
    
\end{itemize}

\section{Background Material}

\subsection{Markov Decision Process}
To enable a reinforcement learning (RL) approach for the trajectory repair problem, we formulate it within the standard framework of a CSA-MDP ~\cite{sutton2018reinforcement},
defined as the tuple
$\langle \mathrm{S}, \mathrm{A}, \mathrm{P}, \mathrm{R}, \gamma \rangle$, where
$\mathrm{S}$ denotes the state space that may be infinite, and the action space 
$\mathrm{A} \subseteq \mathbb{R}^n$ is continuous.
The transition dynamics are described by a probability density
$p(s' | s, a)$, and the reward function is
$\mathrm{R} : \mathrm{S} \times \mathrm{A} \rightarrow \mathrm{R}$.
\noindent An agent interacts with the environment through a stochastic policy 
$\pi(a \mid s)$, which induces a distribution over actions given a state.
The objective in RL is to find a policy $\pi^*$ that
maximizes the expected discounted return:
\[
J(\pi)
=
\mathbb{E}_{\pi}
\!\left[
    \sum_{t=0}^{\infty}
    \gamma^t\, \mathrm{R}(s_t, a_t)
    \,\Big|\, s_0 = s
\right].
\]

\subsection{Proximal Policy Optimization}
In continuous-action settings, policy-gradient methods such as Proximal Policy Optimization (PPO) \cite{schulman2017proximal} are widely used due to their stability and scalability. PPO iteratively updates the policy to maximize $J(\pi)$ while constraining the deviation from previous policies via a clipped surrogate objective. Let $\pi_\theta$ denote the current policy and $\pi_{\theta_{\text{old}}}$ the policy before the update. The probability ratio is defined as:
{
\begin{equation}
    r_t(\theta) = \frac{\pi_\theta(a_t \mid s_t)}{\pi_{\theta_{\text{old}}}(a_t \mid s_t)}.
\end{equation}
}

\noindent The clipped surrogate objective is given by:
{
\begin{equation}
\label{eq:ppo_objective}
    L^{\text{CLIP}}(\theta) = \mathbb{E}_t \left[
    \min\left(
        r_t(\theta)\hat{A}_t,\;
        \mathrm{clip}\big(r_t(\theta), 1 - \epsilon, 1 + \epsilon\big)\hat{A}_t
    \right)
    \right],
\end{equation}
}
\noindent where $\hat{A}_t$ denotes the advantage estimate and $\epsilon$ is a hyperparameter controlling the trust region. In this work, we employ PPO to optimize the plan-refinement agent, enabling consistent learning within the continuous action space of our model.

\subsection{Optimal Control}
For single-trajectory verification and reward engineering, we consider a standard optimal control formulation in continuous time. The objective is to minimize a performance criterion subject to system dynamics, as well as control and state constraints, with given initial and terminal conditions. While our focus will be on solving the minimum-time problem, we present the general form of the optimization problem here for completeness:
\begin{equation} \label{eq:oc}
\begin{aligned}
& \underset{u, t_f}{\text{minimize}} & & \int_{t_0}^{t_f}{l\big(x,u \big) dt} \\
& \text{subject to}  \\
& & & \dot x(t) = f \big(x,u \big) \\
& & & x(t_0)=x_0 \\
& & & x(t_f)=x_f \\
& & & C_u u(t)  \leq U && t \geq t_0 \\
& & & Cx(t)  \leq X  && t \geq t_0.
\end{aligned}
\end{equation}
Here, $x(t) \in \mathbb{R}^n$ is the state vector, $\dot{x}(t) \in \mathbb{R}^n$ is the derivative of the state with respect to time, and $u(t) \in \mathbb{R}^m$ is the control input to the system. The function $f \big(x,u \big)$ is the system dynamics, and $U$ and $X$ give the component-wise constraints on the control and state vector, respectively.

\section{Problem Formulation}

\subsection{Hybrid Planning Problem} 

Hybrid planning concerns decision problems that combine discrete choices with the evolution of continuous system variables. Since our refinement pipeline begins with plans generated by a hybrid planner, we adopt a problem specification consistent with the formalism of such plans. However, since we model a second-order system where acceleration is directly controlled (in contrast to velocity-controlled first-order models), we introduce the following modifications:

\begin{itemize}\setlength\itemsep{0em}
    \item $P$ is the set of propositions that defines the discrete part of the system.
    \item $V = \langle X, X_C \rangle$ where $X$ is the set of real-valued continuous variables that are governed by a set of first-order differential equations denoted as $\dot{X} = \{\dot{x}_1,\dots,\dot{x}_n\}$, and $X_C$ is the set of constraints operating on these variables.
    \item $A$ is the set of durative activities. Each activity is associated with two time points: its start and its end. Each time point can have both discrete and continuous conditions that must be met simultaneously at that exact time point. Additionally, each action has an invariant condition that must hold throughout its execution, and continuous effects that modify the values of continuous variables $X$ during its execution.
    
    \item $I = \langle x_0,\, p_0 \rangle$ is the initial state, a complete assignment over the propositional variables $p_0=p(0)$ and state variables $x_0=x(0)$ at the beginning of the process.

    \item $G = \langle X_G,\, P_G \rangle$ is the goal set specifications, consisting of
    continuous end-of-plan constraints $X_G$ and propositional goals $P_G$ which have to be satisfied at the end of the plan. As such, $G$ is a partial assignment of $P$ and $X$.

    \item $C = \langle U,\, U_C \rangle$ where $U$ is the vector of control variables (accelerations), and $U_C$ is the set of constraints operating on these variables.

    \item $J$ denotes the optimization criterion, which in this work is the
    makespan (total execution time) of the plan.
\end{itemize}

\subsection{Definitions}

\smallskip\noindent\textbf{Definition 1} (Event). An event is the term used to describe the switch between control modes. Events correspond to the start and end points of durative activities.

\smallskip\noindent\textbf{Definition 2} (Hybrid plan skeleton). The plan skeleton $s$ is the ordered list of events $s = (e_0,e_1,\dots,e_n)$, which are the start and end happenings of the durative actions. Note that the plan skeleton is merely the order of events without the assignment of all continuous variables (including timers), which are not concurrent, and separated by at least an $\epsilon$ time constant~\cite{fox2003pddl2}. A plan skeleton and assignment of all the continuous variables and durations (timers) is a hybrid grounded plan.

\smallskip\noindent\textbf{Definition 3} (Continuous Assignment).
A \textit{continuous assignment} $m$ specifies the continuous control inputs associated with each event in the hybrid plan skeleton.
Given a skeleton $s = (e_0, e_1, \dots, e_n)$, the continuous assignment is defined as $m = \{\, m(e_0),\, m(e_1),\, \dots,\, m(e_n) \,\}$
where $m(e_i)$ denotes the continuous assignment to the state variables in $e_i$.

\smallskip\noindent\textbf{Definition 4} (hybrid grounded plan). \textit{A hybrid grounded plan} is defined as a tuple $\langle\mathcal{T},u_t:\mathbb{R}\rightarrow\mathbb{R}^m\rangle$ where:

\begin{itemize}\setlength\itemsep{0em}
    \item $\mathcal{T}$ is the activity schedule, which denotes when each activity should start. $\mathcal{T}$ is given by a list of triplets $\langle a, \tau_s, d\rangle$ where $a$ is an activity, $\tau_s$ the activity start time, and $d$ is the activity duration.
    \item $u_t:[0,T] \rightarrow \mathbb{R}^m$ is the control trajectory, which assigns values to all the inputs (control variables) at every time point between $0$ and $T$.
\end{itemize}

\noindent A valid hybrid grounded plan is a hybrid plan that satisfies all the constraints of the problem, defined by the bounds on the continuous components of the system and the preconditions of the durative actions, and reaches the goal set.

\subsection{Constraint Refinement}
We formulate the refinement problem as a constrained optimization problem built upon the optimal control formulation proposed in \cite{taitler2019minimum}. Given a hybrid grounded plan produced by an off-the-shelf hybrid planner, let $s = (e_0, e_1, \dots, e_n)$ denote its plan skeleton and $(t_0, t_1, \dots, t_n)$ the corresponding timestamps. Each consecutive pair of events induces a transition interval
\[
I_i = [t_i,\, t_{i+1}], \qquad i \in \{0,\dots,n-1\},
\]
with duration $\Delta t_i = t_{i+1} - t_i$.
\smallskip
For every interval $I_i$, MTV computes the minimum physically achievable
traversal time $t_{\min,i}$ under the non-linear second-order
dynamics
\begin{equation} \label{eq:second-order}
    \begin{aligned}
        \dot{x}_1(t) &= x_2(t), \\[3pt]
        \dot{x}_2(t) &= u(t) - \tfrac{1}{2}k\,x_2^2(t), \\[3pt]
    \end{aligned}
\end{equation}
where $x=[x_1,x_2]^\top$ is the state vector, composed of the position and velocity respectively, $u(t)$ is the acceleration control input, and $k$ is a drag coefficient.
A hybrid grounded plan is therefore
\emph{physically feasible} if
\[
\Delta t_i \ge t_{\min,i}
\qquad \forall i \in \{0,\dots,n-1\}.
\]

In the refinement setting considered in this work, the discrete plan
skeleton $s$ remains fixed and only the continuous assignment is
modified. Let $\mathrm{M}(s)$ denote the set of all continuous assignments
compatible with the event sequence $s$. Each $m \in \mathrm{M}(s)$ induces
timestamps $(t_0, t_1, \dots, t_n)$ and corresponding interval durations
$\{\Delta t_i(m)\}$. The refinement problem can therefore be formulated as the following
constrained optimization problem:
\[
\begin{aligned}
\min_{m \in \mathrm{M}(s)} \quad & J(m) \\
\text{s.t.} \quad
& \Delta t_i(m) \ge t_{\min,i}(m),
\quad \forall i \in \{0,\dots,n-1\}.
\end{aligned}
\]
Where $J$ is the objective function that the off-the-shelf hybrid planner is designed to minimize. The aim is to compute a continuous assignment for the fixed hybrid
plan skeleton such that all intervals satisfy the MTV-derived second-order feasibility bounds, while preserving the discrete-event structure and limiting the increase in makespan $T$.

\section{Methodology}
\begin{figure}[t!]
    \centering
    \includegraphics[width=1.0\linewidth]{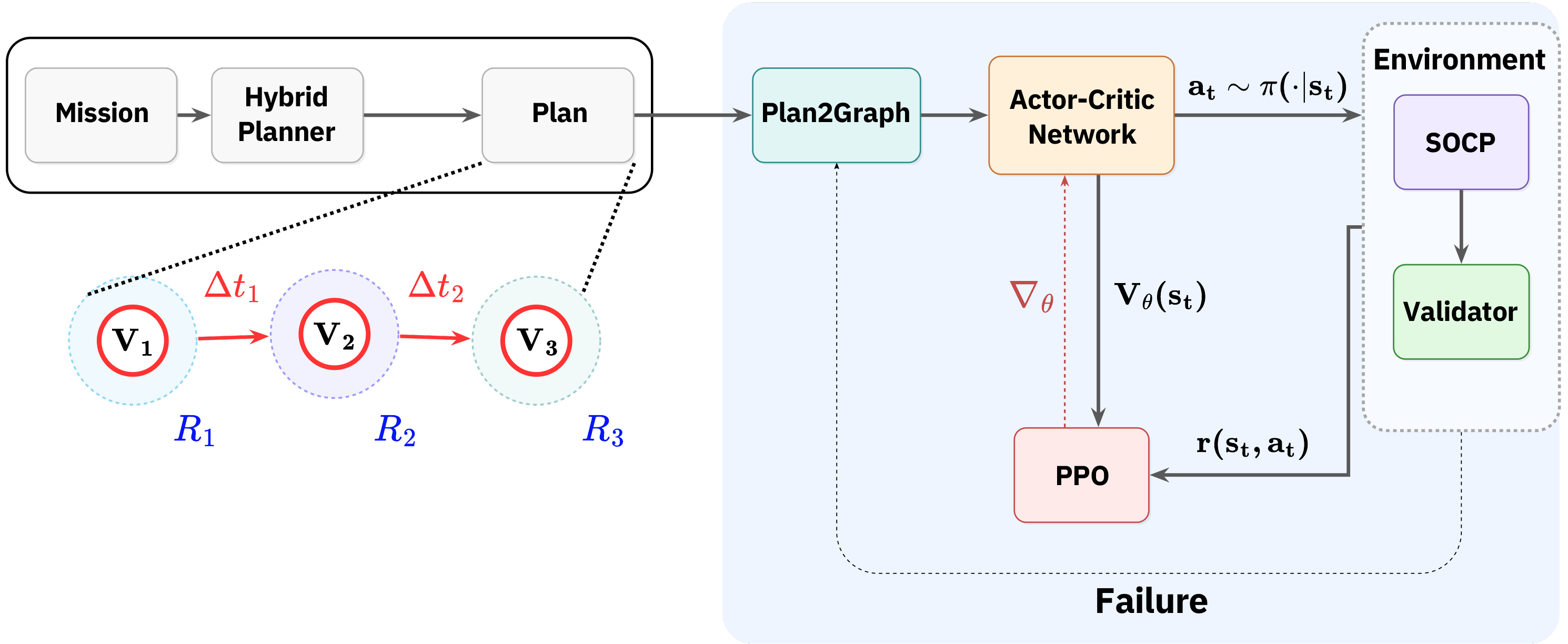}
    \caption{The refinement framework. A hybrid planner generates a first-order plan, which is converted to a graph. A GNN-based RL agent adjusts velocity bounds, and an SOCP solver generates a candidate trajectory validated by MTV.}
    \label{fig:pipline}
\end{figure}

We address the refinement problem introduced in the previous section by constructing a learning-based framework whose goal is to find a continuous assignment for the fixed hybrid plan skeleton that induces a physically feasible hybrid grounded plan that is also makespan-optimized. The refinement problem is cast as a CSA-MDP. The state is the graph $G=(V,E)$ describing the current plan. The policy outputs continuous scaling factors that adjust velocity bounds for all motion segments simultaneously. After applying the action, an SOCP solver, that is formulated as the one in \cite{fernandez2018scottyactivity}, minimizes the plan makespan subject to the new bounds and original temporal constraints. The resulting trajectory is evaluated using MTV to compute a reward. We employ RL, specifically PPO, because the optimization landscape defined by the interaction between velocity bounds, the SOCP solver, and the closed-form MTV validation is highly non-linear and non-differentiable. Specifically, the objective's gradient is discontinuous at the switching surfaces between the optimal control regimes, making stochastic policy gradients a suitable choice for navigating this non-convex space without subgradient calculus. The final output comprises the refined plan and the valid second-order MTV motion profiles. The refinement process consists of four stages (Fig.~\ref{fig:pipline}). (I) We obtain a first-order-feasible hybrid grounded plan from a hybrid planner; (II) We convert this plan into a structured graph that captures the spatial, temporal, and dynamical relationships while respecting the fixed event sequence; (III) Each transition interval is evaluated using the closed-form MTV computation to identify violations of the second-order feasibility constraints.
Finally, (IV) a reinforcement-learning agent iteratively proposes adjustments to velocity bounds; each is transformed into a candidate
trajectory via an SOCP solver and validated again by MTV. This iterative loop terminates after a pre-defined finite horizon. To preserve MDP stationarity without explicitly encoding the remaining timestep in the state, the agent optimizes an infinite-horizon discounted objective.

\subsection{Plan-to-Graph Representation}

The hybrid grounded plan returned by the hybrid planner is transformed into a graph
$G = (V,E)$ that serves as the state representation for our MDP formulation. Each event
$e_i$ becomes a node $v_i \in V$ with attributes capturing its position,
timestamp, dwell interval, and regional memberships.

\smallskip\noindent\textbf{Edge Construction.}
Every two consecutive events $(e_i, e_{i+1})$ are mapped to a directed edge \((v_i, v_{i+1}) \in E\) encoding the continuous constraints governing the transition between these events, including the segment duration, the velocity bounds assigned by the planner, and the maximum-norm constraint. 
Consecutive events with negligible spatial displacement, i.e., a durative activity that constrains the movement to remain static,
are merged to avoid near-zero-length segments; durations are stored as a dwell-time attribute of the node $v_{i+1}$.

\smallskip\noindent\textbf{Node Velocity Estimation.}
As first-order planners treat velocity as a control variable, node
velocities are not directly provided. To fill in the missing values and obtain a second-order state
representation, we estimate the velocity at each intermediate node $v_{i} \in V$ by
taking a duration-weighted average of its incoming and outgoing segment
velocities. Let $\Delta t_{i-1}$ and $\Delta t_{i}$ denote the durations of the consecutive events
$(v_{i-1},v_{i})$ and $(v_{i},v_{i+1})$, and let $v_x(\cdot)$ and
$v_y(\cdot)$ be their velocities along each axis. Then

{\small
\[
v_{i,v_j} =
\frac{\Delta t_{i-1}\, v_j(v_{i-1},v_{i}) +
      \Delta t_{i}\, v_j(v_{i},v_{i+1})}
     {\Delta t_{i-1} + \Delta t_{i}}, \quad j \in \{x,y\}.
\]
}

\noindent For the terminal node, we set $v_{v_x} = 0$ and $v_{v_y} = 0$. To preserve the
direction of motion along the trajectory, we determine the sign of each velocity
component based on the displacement between consecutive nodes. Specifically, for
each intermediate node $v_i$, we compute the differences between
$(v_{i,x}, v_{i,y})$ and $(v_{i+1,x}, v_{i+1,y})$, and assign the final velocities as
$v_{i,v_x} = \mathrm{sign}(v_{i+1,x} - v_{i,x}) \cdot v_{i,v_x}$ and
$v_{i,v_y} = \mathrm{sign}(v_{i+1,y} - v_{i,y}) \cdot v_{i,v_y}$.

\smallskip\noindent\textbf{Graph-based State Representation.}
The graph is processed by a GNN that encodes spatial, temporal, and regional information into latent embeddings for the policy and value function. Each node \(v \in V\) is represented by 7 scalar features and a region embedding:
\[
\mathbf{h}_v =
(x, y, v_x, v_y,
t,\; \tau_{\text{low}}, \tau_{\text{high}},\; \mathbf{r}_{\text{enc}}),
\]
where $\tau_{\text{low}}$ and $\tau_{\text{high}}$ denote the lower and upper bounds on
the dwell time, respectively, and $\mathbf{r}_{\text{enc}}$ denotes a learned region
embedding. Region information is inherently heterogeneous; to enable uniform processing, all
regions are encoded using a unified representation scheme. Circles and rectangles
are parameterized as $[c_x, c_y, r_x, r_y, \mathbf{1}_{\text{type}}]$, while polygons
are encoded using a Deep Sets~\cite{zaheer2017deep} architecture to ensure permutation
invariance across vertices. When multiple regions of the same type are present, their embeddings are averaged. When regions of different types are present, their embeddings are concatenated and passed through an intersection layer, ensuring that the resulting representation satisfies the requirement that each trajectory point satisfies all associated region constraints.
\noindent Each edge $(v_i,v_{i+1})\in E$ carries:
\[
\mathbf{h}_{(v_i,v_{i+1})} =
\begin{aligned}
\big(
& v_x^{\min},\, v_x^{\max},\;
  v_y^{\min},\, v_y^{\max}, b_{norm}, d_{\min},\, d_{\max}
\big)
\end{aligned}
\]
Here, $v_x^{\min},\, v_x^{\max},\;
  v_y^{\min},\, v_y^{\max}$ are per-axis velocity bounds, $b_{norm}$ is the maximum velocity norm bound, and $d_{\min},\, d_{\max}$ are the duration bounds of the durative activity.

\subsection{Minimum-Time Validation}

MTV is a feasibility analysis method designed to verify whether a continuous transition between two states can be executed under non-linear second-order dynamics~\eqref{eq:second-order} with bounded acceleration, bounded velocity, and quadratic drag~\cite{taitler2019minimum}. A state is defined to be a vector of $[x_1, x_2]^\top$
where $x_{1}$ represents the position and $x_{2}$ represents the velocity of the agent. The problem is formulated as an optimal control problem as in \eqref{eq:oc}, specifically a minimum-time optimal control problem defined as follows:

\begin{equation} \label{eq:second-order-optimization}
\begin{aligned}
    & \underset{u, t_f}{\text{minimize}} & & \int_0^{t_f}1\cdot dt \\
    & \text{subject to}  \\
    &&&\dot{x}_1(t) = x_2(t), \\[3pt]
    &&&\dot{x}_2(t) = u(t) - \tfrac{1}{2}k\,x_2^2(t), \\[3pt]
    &&&|u(t)| \le U, \quad |x_2(t)| \le V, \quad t \ge t_0, \\[3pt]
    &&&x_1(t_0) = x_{10}, \quad x_2(t_0) = x_{20}, \\[3pt]
    &&&x_1(t_f) = x_{1f}, \quad x_2(t_f) = x_{2f}.
\end{aligned}
\end{equation}

\noindent Here, $U$ is the maximum acceleration, $V$ the maximum allowable velocity,
and $k$ is a drag coefficient. Given the initial $[x_{10}, x_{20}]^\top$
and terminal states $[x_{1f},x_{2f}]^\top$,
MTV computes the
minimum physically achievable traversal time $t_{\min}$ for the segment. Intuitively, $t_{\min}$ represents the fastest motion permitted by the agent's
acceleration, velocity, and drag limits. If the planner assigns a duration
$d < t_{\min}$, the segment is physically impossible, even though it may appear
consistent under a first-order model. MTV admits closed-form expressions for $t_{\min}$, depending on whether the velocity bound $V$ is reached. 
The resulting motion is either Bang-Constant-Bang or Bang-Bang.

\smallskip\noindent\textbf{Bang-Constant-Bang (BCB).}
When the optimal motion reaches the velocity bound $V$, the solution takes a
Bang-Constant-Bang (accelerate-cruise-decelerate) form. This occurs when

\begin{equation}
\begin{aligned}
    V &< \sqrt{\tfrac{2U}{k}}, \qquad
    \Delta x_a + \Delta x_d < x_{1f} - x_{10}, \\[1pt]
    \Delta x_a &= \tfrac{1}{k}\ln\!\left(
        \tfrac{2kU - (x_{20}k)^2}{2kU - (Vk)^2}
    \right), \\[1pt]
    \Delta x_d &= \tfrac{2}{k}\ln\!\left(
        \tfrac{2kU + (Vk)^2}{
            \sqrt{(2kU + k^2 V x_{2f})^2 + 2k^3U(V - x_{2f})^2}
        }
    \right).
\end{aligned}
\label{eq:profile-conditions}
\end{equation}

The segment duration is the sum of three phases:

\begin{equation}
\begin{aligned}
    \Delta t_1 &= \tfrac{1}{\sqrt{2kU}}
    \ln\!\left(
        \tfrac{ \sqrt{2kU} - x_{20}k }{ \sqrt{2kU} + x_{20}k } 
        \cdot 
        \tfrac{ \sqrt{2kU} + Vk }{ \sqrt{2kU} - Vk }
    \right), \\[1pt]
    \Delta t_2 &= 
    \tfrac{x_{1f} - x_{10} - \Delta x_a - \Delta x_d}{V}, \\[1pt]
    \Delta t_3 &= 
    \sqrt{\tfrac{2}{kU}}\,
    \tan^{-1}\!\left(
        \tfrac{ \sqrt{2k^3U}(V - x_{2f}) }{ 2kU + x_{2f}k^2V }
    \right), \\[1pt]
    t_{\min} &= \Delta t_1 + \Delta t_2 + \Delta t_3.
\end{aligned}
\label{eq:BCB-minimum-time}
\end{equation}

\smallskip\noindent\textbf{Bang-Bang (BB).}
If the velocity bound is not reached (the conditions in (\ref{eq:profile-conditions}) do not hold), the optimal control is Bang-Bang
(accelerate-decelerate).  
The peak velocity $\tilde{V}$ is the positive root of the quadratic

\begin{equation}
\begin{aligned}
    a\tilde{V}^4 + b\tilde{V}^2 + c &= 0, 
    \qquad 
    \tilde{V} = 
    \sqrt{\tfrac{-b + \sqrt{b^2 - 4ac}}{2a}},
\end{aligned}
\end{equation}

and the minimum time is

\begin{equation}
\begin{aligned}
    t_{\min} &= \tfrac{1}{\sqrt{2kU}}
    \ln\!\left(
        \tfrac{ \sqrt{2kU} - x_{20}k }{ \sqrt{2kU} + x_{20}k } 
        \cdot 
        \tfrac{ \sqrt{2kU} + \tilde{V}k }{ \sqrt{2kU} - \tilde{V}k }
    \right) \\[3pt]
    &\quad + 
    \sqrt{\tfrac{2}{kU}}\,
    \tan^{-1}\!\left(
        \tfrac{\sqrt{2k^3U}(\tilde{V} - x_{2f})}{
              2kU + x_{2f}k^2\tilde{V}}
    \right).
\end{aligned}
\label{eq:BB-minimum-time}
\end{equation}

\smallskip\noindent\textbf{Extension to 2D.}
Following the standard assumptions adopted by common hybrid planners \cite{fernandez2018scottyactivity}, we model the two axes as independent. Since no external coupling forces are considered \cite{taitler2022time}, the two-dimensional validation problem decomposes into two separate one-dimensional MTV computations. The motion duration is then determined by the axis requiring the longer traversal time. The shorter axis is automatically constrained by the longer axis, resulting in a BCB profile by definition. A segment is considered physically feasible only if both axes individually satisfy their respective dynamic bounds:
\[
t_{\min}^{2D} = \max \{ t_{\min}^x,\; t_{\min}^y \}.
\]

\noindent Analytically, this decoupling constructs a conservative inner approximation of the dynamically feasible control space, guaranteeing strict satisfaction of the maximum-norm control constraints ($||u(t)|| \le U$). While MTV is a powerful feasibility checker, it is strictly a validator: it cannot modify or repair a hybrid plan that violates second-order dynamics. In our framework, MTV quantifies the feasibility gap between the first-order hybrid plan and the true physical dynamics; this gap serves as the learning signal for our reinforcement-learning refinement method.

\subsection{CSA-MDP Formulation}

The refinement loop is modeled as a CSA-MDP:
\[
\langle \mathrm{S},\mathrm{A},\mathrm{P},\mathrm{R},\gamma\rangle
\]
\noindent\textbf{State space.}
    Each state $s\in\mathrm{S}$ is the complete assignment of the continuous variables at the event points of the plan skeleton represented by a graph $G=(V,E)$ produced by the
    plan-to-graph conversion.
    
\noindent\textbf{Action space.} 
The space of possible changes to the velocity bounds between the plan skeleton events is represented by a continuous matrix $\mathbf{a_t}\in[0,1]^{|E|\times 2}$ produced by a sigmoid-squashed Gaussian policy:
\[
    \mathbf{a_t} \mid s_t \sim \text{sigmoid}(\mathcal{N}(\mu(s_t),\sigma(s_t))).
\]
Each row $\mathbf{a_{t,i}} = (a_{t,i,x},a_{t,i,y}) \in [0,1]^2$, associated with edge $i \in E$, specifies independent refinements to the velocity bounds along the $x$- and $y$-axes. Each entry of $a_{t,i}$ represents the percentage of decrease in velocity per axis. While this strictly monotonic contraction forces a conservative inner-approximation search, the parameterized GNN policy amortizes this optimization cost by generalizing across the training distribution. For each edge $i \in E$, let $v_{x,t,\max}^{(i)}$ and $v_{y,t,\max}^{(i)}$ denote its current velocity bounds at time $t$. Then, for each action component $a_{t,i,j} \in [0,1]$, the corresponding velocity bounds are updated according to

\[
v_{j,t+1,\max}^{(i)} = a_{t,i,j}*v_{j,t,\max}^{(i)},  \quad v_{j,t+1,\min}^{(i)} = -v_{j,t+1,\max}^{(i)}.
\]

\noindent \textbf{Transition dynamics.}
The transition function is defined as follows: if the SOCP solver returns a first-order feasible solution, the system transitions from $s_t$ to a new state $s_{t+1}$. Otherwise, the system transitions to a terminal state $s_T$. This occurs when the RL agent has reduced the velocity beyond a point where further reductions no longer affect SOCP feasibility.

\noindent\textbf{Reward Function.}
The reward is defined based on the feasibility of the solution returned by the SOCP solver. If the SOCP solver fails to produce a feasible solution, no new continuous assignment is generated, the environment transitions to a terminal state, and the agent receives a reward of $-1$. If the SOCP solver produces a first-order feasible solution, a new continuous assignment $m$ is associated with the hybrid plan skeleton. For each edge $i$, we evaluate second-order feasibility using MTV. Let $\Delta t_i$ denote the duration assigned by the SOCP for edge $i$, and let $t_{\min}^{(2)}(i)$ denote the minimum traversal time under second-order dynamics. We define the \emph{feasibility gap} as
\[
g_i = \min\!\left(0,\; \Delta t_i - t_{\min}^{(2)}(i)\right),
\]
so that negative values indicate second-order violations. The corresponding per-edge reward is $r_i = \frac{g_i}{t_{\min}^{(2)}(i)}$ and the final immediate reward is $r_{gap} = \sum_{i\in E} \frac{1}{|E|}r_i$. The agent receives this feasibility-gap reward only when the solution is first-order feasible but violates second-order dynamics. If the solution is also second-order feasible, meaning $g_i = 0, \forall~i\in E$, the reward instead reflects the change in makespan relative to the initial plan. Let $T_{\text{planner}}$ denote the makespan of the original hybrid plan and $T_{\text{SOCP}}$ the makespan after refinement. The reward in this case is defined as $\frac{T_{\text{planner}}}{T_{\text{SOCP}}}$. Note that the feasibility-gap reward is non-positive, whereas the makespan-based reward is positive. Although this introduces a structural discontinuity exactly at the feasibility boundary, the stochasticity of the policy smoothly approximates the expected return $J(\pi)$ for reliable policy gradient estimation. The final immediate reward per step function is then:

\begin{align}
r_t &=
\begin{cases}
-1, & \text{if SOCP is infeasible}, \\[6pt]
r_{\text{gap}}, 
& \text{if SOCP is feasible and}~ \exists i \in E:\; g_i < 0, \\[6pt]
\displaystyle \frac{T_{\text{planner}}}{T_{\text{SOCP}}}, 
& \text{if } g_i = 0,\ \forall i \in E.
\end{cases}
\end{align}

\subsection{Training Procedure}
In a continuous control problem, a known issue is how to slow down and constrain the learning to the current knowledge already obtained as learning progresses. This has been studied in various RL algorithms \cite{taitler2017learning, farsang2021decaying}, and here we employ clip decay as the suitable remedy when using PPO \cite{farsang2021decaying}. The clip ratio decay used in this work follows the linear schedule proposed by~\cite{farsang2021decaying}. Let $K$ denote the total number of updates, $k$ the current update, and $\epsilon_0$ the initial PPO clip ratio. The decayed clip ratio is defined as:
\[
\epsilon_k = \frac{K - k}{K}\,\epsilon_0.
\]
Since this schedule reduces the clip ratio from $\epsilon_0$ to $0$, which may overly restrict learning in later stages, we introduce a minimum bound $\epsilon_{\min}$. The final update rule is therefore:
\[
\epsilon_k = \max\!\left(\epsilon_{\min},\; \frac{K - k}{K}\,\epsilon_0 \right).
\]

\begin{figure}[t!]
    \centering
    \includegraphics[width=\linewidth]{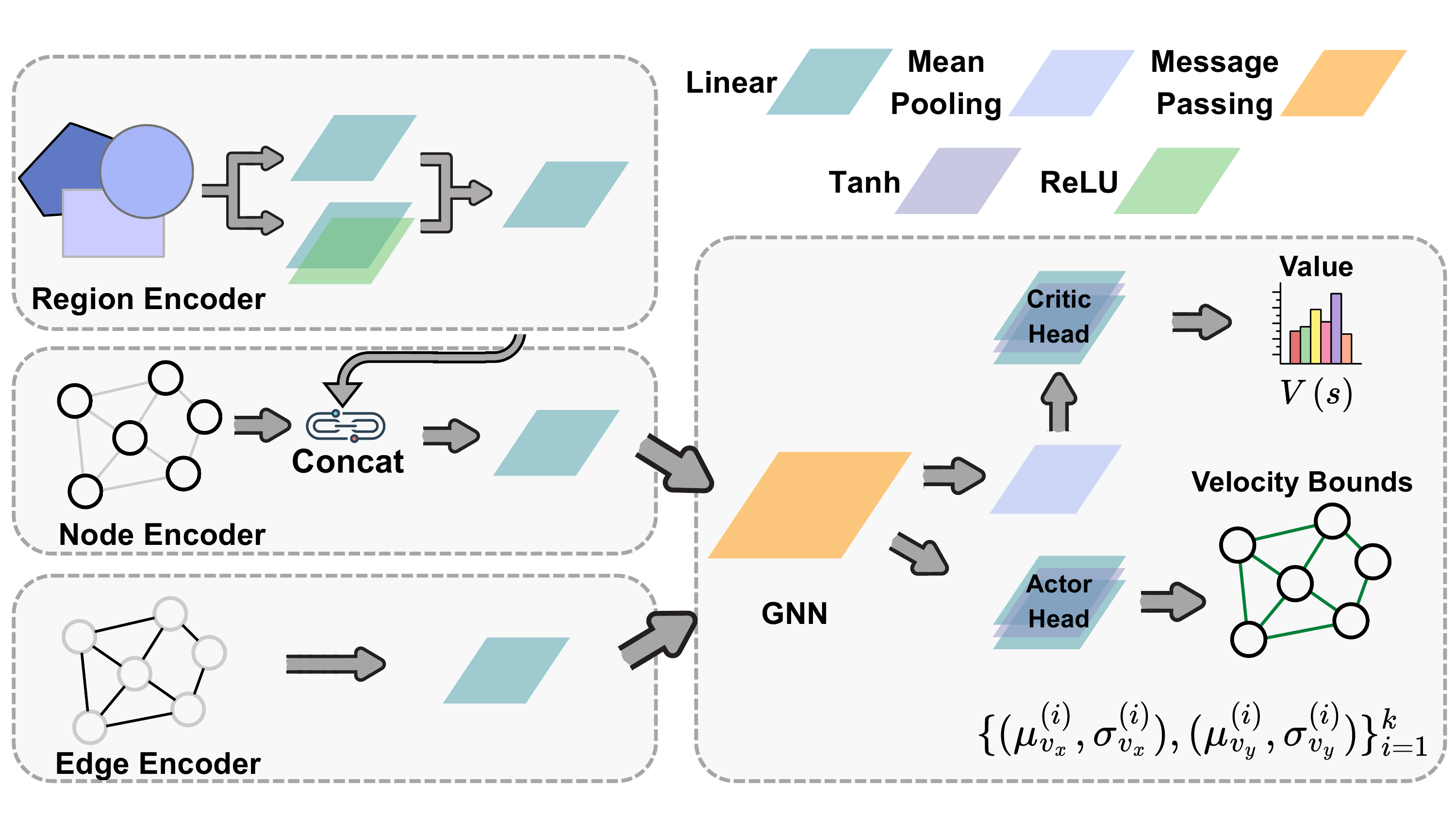}
    \caption{Shared-encoder actor-critic architecture. Region, node, and edge features are encoded separately and concatenated before being processed by a message-passing GNN. The resulting node embeddings are used by separate actor and critic heads to produce Gaussian policy parameters and value estimates.}
    \label{fig:architecture}
\end{figure}

All models were trained using a shared-encoder actor-critic PPO architecture with graph-based state encoding across 5 random seeds (see Table~\ref{tab:ppo_hparams} for full hyperparameters). The network consists of three input encoders, a shared GNN, and separate actor and critic heads. The Region Encoder processes geometric features via parallel $\text{Linear}(5,16)$ and $\text{Linear}(2N,16)+\text{ReLU}$ layers, followed by $\text{Linear}(16,16)$ when multiple region types are present. The Node Encoder concatenates node and region features and maps them through $\text{Linear}(23,32)$, while the Edge Encoder applies $\text{Linear}(9,32)$ to edge features. These representations are concatenated and passed through a GNN with two message-passing layers, $\text{Linear}(96,32)+\text{Tanh}$ and $\text{Linear}(64,32)+\text{Tanh}$, producing 32-dimensional node embeddings. The critic head applies mean pooling followed by $\text{Linear}(32,32)+\text{Tanh}+\text{Linear}(32,1)$, while the actor head uses $\text{Linear}(64,32)+\text{Tanh}+\text{Linear}(32,4)$ to output Gaussian policy parameters (see Fig.~\ref{fig:architecture}). Both heads use orthogonal initialization.

\begin{table}[t!]
\caption{PPO Hyperparameters}
\label{tab:ppo_hparams}
\centering
\footnotesize 
\renewcommand{\arraystretch}{1.05} 
\setlength{\tabcolsep}{3.5pt} 

\begin{tabularx}{\columnwidth}{l X l X}
\toprule
\textbf{Hyperparameter} & \textbf{Value} & \textbf{Hyperparameter} & \textbf{Value} \\
\midrule
Learning Rate & $10^{-4}$ & Target KL Div & 0.015 \\
Clip Ratio ($\epsilon_0$) & 0.2 & Discount ($\gamma$) & 0.99 \\
Min Clip Ratio ($\epsilon_{\min}$) & 0.05 & GAE $\lambda$ & 0.95 \\
Batch Size & 16 & Horizon & 8 \\
Entropy Coeff. & 0.01 & Rollout Size & 64 \\
Value Loss Coeff. & 0.5 & Epochs & 1 \\
Max Grad Norm & 0.5 & Episodes & 15{,}000 \\
Optimizer & Adam & Seeds & 1,2,3,4,5 \\
\bottomrule
\end{tabularx}
\end{table}

\section{Experiments}
We evaluate our method across four hybrid planning domains. Three domains were originally used in the evaluation of Scotty~\cite{fernandez2018scottyactivity}, and one domain originates from the 2023 International Planning Competition's numeric track~\cite{taitler2024ipc}.

Each training job was allocated an NVIDIA L40 GPU, 3~vCPUs, and 8\,GiB of RAM. Initial hybrid plans were generated using the off-the-shelf planner Scotty \cite{fernandez2018scottyactivity}. Second-order dynamics parameters were set to $U=10$, $V=10$, and $k=0.05$, in alignment with \cite{taitler2019minimum}. 

\noindent\textbf{AUV-Domains.}
The \textsc{AUV-2D} and \textsc{Norm-AUV-2D} domains model underwater sampling missions in which an autonomous underwater vehicle (AUV) must visit and sample from specified regions. 
The former does not impose maximum-norm velocity constraints, whereas the latter enforces such bounds, resulting in more restrictive dynamics.

\noindent\textbf{OnAir-Refuel.}
This domain models an aerial refueling mission involving a jet and a tanker. Since multi-agent dynamics are outside the scope of this work, we simplify the domain by forcing the tanker to move along a fixed straight-line path between two refueling regions. This yields an equivalent single-agent setting compatible with our framework.

\noindent\textbf{Sailing.}
The \textsc{Sailing} domain, from the 2023 IPC, models a vessel navigating between polygonal regions under wind-dependent motion. 
The domain also includes tight temporal constraints and region-based goals. We use this domain primarily to test refinement on a single-axis motion model while handling polygonal constraints, as motion disturbances are not part of the scope of this work.

\noindent\textbf{Baseline.} We compare our method against a conservative baseline given by a constant contraction policy $\pi(s)=\bar{a}$ where $\bar{a}\in[0,1]$ is applied uniformly at each refinement step to the velocity bounds of all edges and both axes. We evaluate $\bar{a}\in\{0.9,\,0.995\}$, corresponding to per-step reductions of 10\% and 0.5\%, respectively. More aggressive contraction reaches feasibility in fewer iterations but tends to produce more conservative, higher-makespan solutions, whereas finer contraction may yield tighter feasible solutions at the expense of additional iterations.


\begin{figure}[t!]
    \centering
    \includegraphics[width=\linewidth]{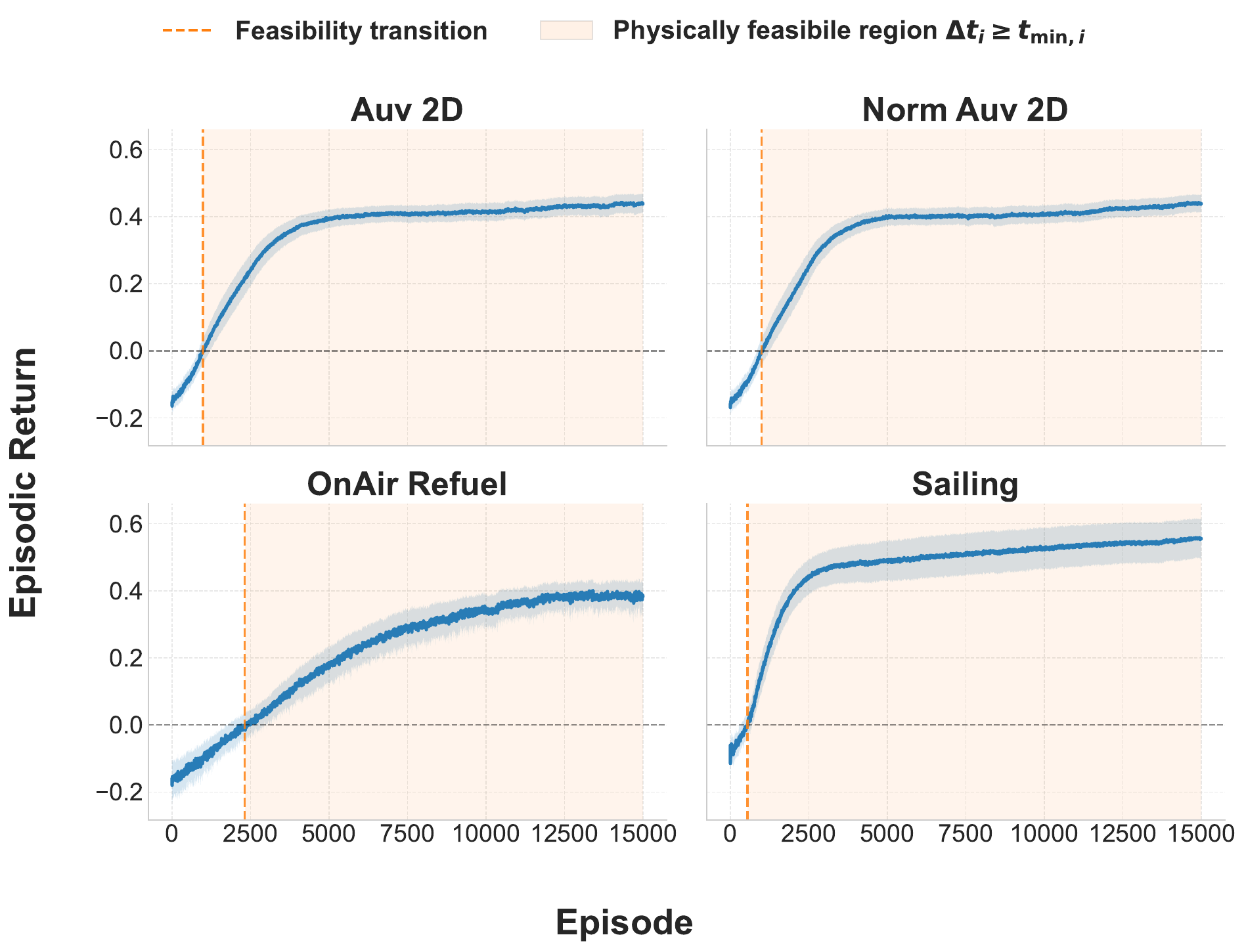}
    \caption{Training curves across domains. The agent consistently improves its return and reaches a stable regime.}
    \label{fig:episode_return_by_domain}
\end{figure}

\subsection{Results}
Table~\ref{tab:makespan_ratio_domains} summarizes the performance of the refined solutions. Enforcing second-order dynamics renders all initial plans from the first-order planner physically infeasible (0\% success). In contrast, our refinement framework recovers feasibility in 100\% of cases, bridging the gap between symbolic planning and realizable execution. This comes at the cost of increased makespan (ratios $>1$), reflecting the underestimation inherent in first-order models. The resulting makespan thus provides a strict upper bound on the global optimum of the underlying MINLP.


Fig.~\ref{fig:episode_return_by_domain} reports the evolution of the average episodic return as a function of episodes across all evaluated domains. It also marks when the RL agent transitions into the physically feasible region and begins optimizing the hybrid plan's makespan relative to the initial plan. Across all domains, the episodic return shows a clear upward trend, indicating consistent improvement of the refinement policy. In the \textsc{Norm-AUV-2D} and \textsc{AUV-2D} domains, the agent achieves near-monotonic return growth. In the \textsc{Sailing} domain, we observe fast convergence, reaching a stable plateau within the first 2{,}000 episodes. Finally, in the \textsc{OnAir-Refuel} domain, the return increases more gradually as refinement progresses.

\begin{figure}[t!]
    \centering
    \includegraphics[width=0.985\columnwidth]{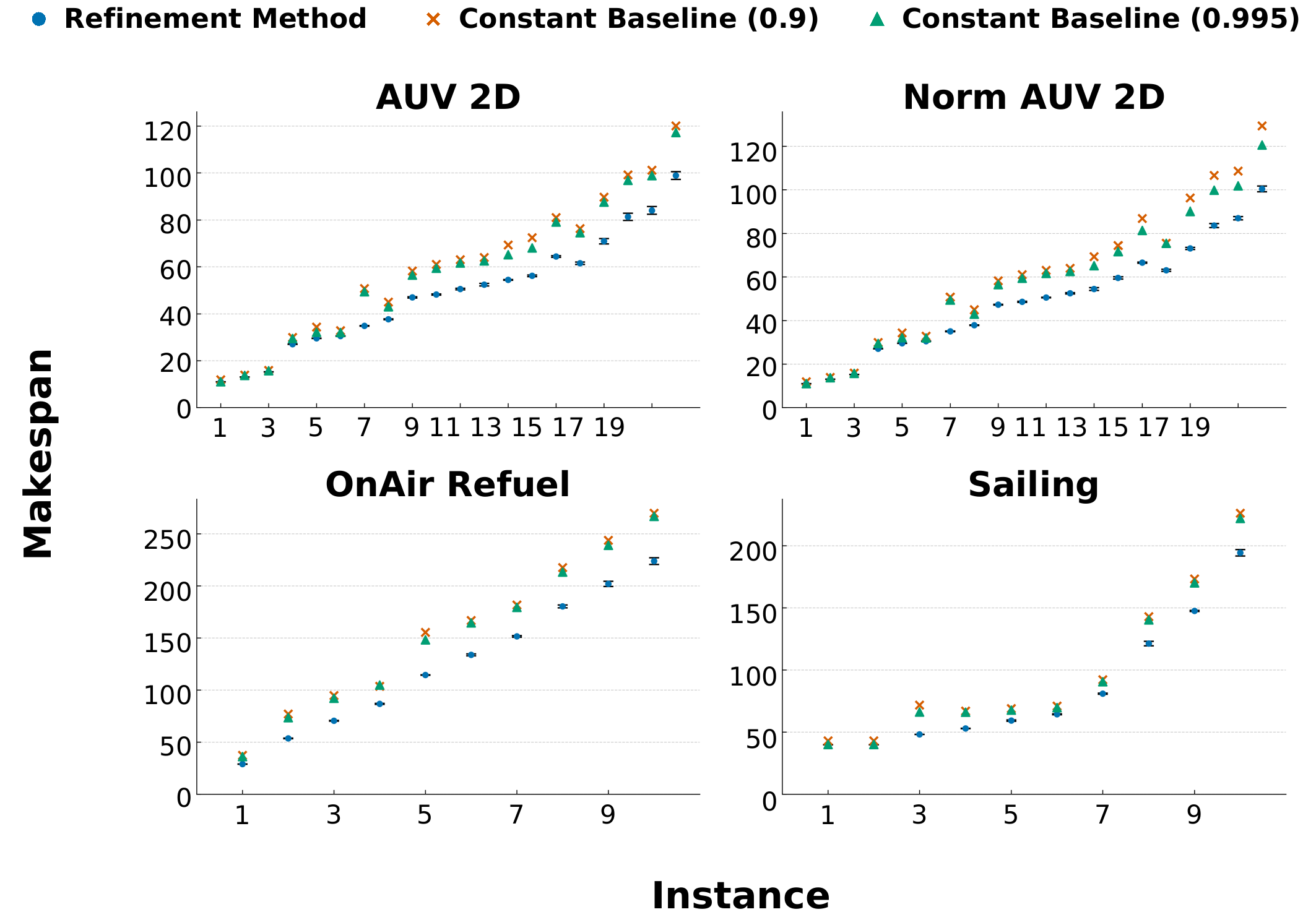}
    \caption{Per-instance makespan comparison. Blue points show
    refined makespan with $\pm 1$ std error bars. Orange and green markers denote the baseline makespan with $\bar{a}=\{0.9,0.995\}$.
    }
    \label{fig:makespan_per_instance_by_domain}
\end{figure}

Fig.~\ref{fig:makespan_per_instance_by_domain} reports the makespan achieved by our refinement method (blue, mean $\pm$ std) compared to constant baselines with uniform velocity contraction factors of 10\% and 0.5\% per step (orange and green, respectively), across all evaluated instances and domains. 
The refined solutions consistently achieve lower makespans than both baselines, demonstrating systematic improvements in execution efficiency.

Notably, the finer-grained baseline (0.5\%) yields only marginal improvement over the coarser 10\% baseline, with an average makespan reduction of just 3.24\% across domains, while requiring, on average, 1784.8\% more refinement steps to reach feasibility. In contrast, our approach leverages a structured MDP formulation to perform segment-wise, physically informed adjustments, yielding more globally optimized, physically consistent solutions.

\begin{table}[tb]
\caption{Performance statistics. We report the number of problem instances and the average makespan ratio of the final refined feasible solution relative to the initial infeasible plan. Values are mean $\pm$ std.\ dev. Initial plan feasibility was 0\% for all domains; refined plan feasibility was 100\%.}
\label{tab:makespan_ratio_domains}
\centering
\footnotesize
\renewcommand{\arraystretch}{1.05}
\setlength{\tabcolsep}{4pt}

\begin{tabular}{l c c}
\toprule
\textbf{Domain} & \textbf{Instances} & \textbf{Makespan Ratio }\\
\midrule
AUV-2D & 20 & $1.20 \pm 0.008$ \\
Norm-AUV-2D & 20 & $1.23 \pm 0.006$ \\
OnAir-Refuel & 10 & $1.15 \pm 0.007$ \\
Sailing & 10 & $1.14 \pm 0.006$ \\
\bottomrule
\end{tabular}
\end{table}



\section{Conclusions and Future Work}
We addressed the gap between hybrid temporal planning under first-order dynamics and the physical feasibility requirements of second-order systems. Empirically, we demonstrated that all initial plans generated by an off-the-shelf hybrid planner are physically infeasible under second-order dynamics, highlighting the severity of the modeling gap. In contrast, our refinement approach recovers feasibility in all cases while maintaining competitive makespan performance, establishing the framework as an effective bridge between symbolic planning outputs and physically valid execution. 
However, the method assumes disturbance-free, decoupled dynamics and no time windows -- assumptions that should be relaxed in realistic settings.
Future work will consider problems with time-window constraints and extend the framework to timing-robust settings with coupled axes and variability in activity durations, further bridging the gap between planning and real-world execution.





\bibliography{aaai2026}

@article{fernandez2018scottyactivity,
  author       = {Enrique Fern{\'{a}}ndez{-}Gonz{\'{a}}lez and
                  Brian C. Williams and
                  Erez Karpas},
  title        = {ScottyActivity: Mixed Discrete-Continuous Planning with Convex Optimization},
  journal      = {Journal of Artificial Intelligence Research},
  volume       = {62},
  pages        = {579--664},
  year         = {2018}
}

@article{fox2003pddl2,
  title   = {{PDDL}2.1: An extension to {PDDL} for expressing temporal planning domains},
  author  = {Fox, Maria and Long, Derek},
  journal = {Journal of Artificial Intelligence Research},
  volume  = {20},
  pages   = {61--124},
  year    = {2003}
}

@article{schulman2017proximal,
  title   = {Proximal policy optimization algorithms},
  author  = {Schulman, John and Wolski, Filip and Dhariwal, Prafulla and Radford, Alec and Klimov, Oleg},
  journal = {arXiv preprint arXiv:1707.06347},
  year    = {2017}
}

@article{zaheer2017deep,
  title   = {Deep sets},
  author  = {Zaheer, Manzil and Kottur, Satwik and Ravanbakhsh, Siamak and Poczos, Barnabas and Salakhutdinov, Russ R and Smola, Alexander J},
  journal = {Advances in neural information processing systems},
  volume  = {30},
  year    = {2017}
}

@book{sutton2018reinforcement,
  title     = {Reinforcement learning: An introduction},
  author    = {Sutton, Richard S and Barto, Andrew G},
  year      = {2018},
  publisher = {MIT press}
}

@article{lobo1998applications,
  title     = {Applications of second-order cone programming},
  author    = {Lobo, Miguel Sousa and Vandenberghe, Lieven and Boyd, Stephen and Lebret, Herv{\'e}},
  journal   = {Linear algebra and its applications},
  volume    = {284},
  number    = {1-3},
  pages     = {193--228},
  year      = {1998},
  publisher = {Elsevier}
}

@inproceedings{taitler2019minimum,
  title={Minimum time optimal control of second order system with quadratic drag and state constraints},
  author={Taitler, Ayal and Ioslovich, Ilya and Karpas, Erez and Gutman, Per-Olof},
  booktitle={{Conference on Decision and Control}},
  pages={523--528},
  year={2019},
  organization={IEEE}
}

@article{taitler2022time,
title = {Time optimal control of a non-linear surface vehicle subject to disturbances},
journal = {IFAC Journal of Systems and Control},
volume = {21},
pages = {100195},
year = {2022},
issn = {2468-6018},
doi = {https://doi.org/10.1016/j.ifacsc.2022.100195},
url = {https://www.sciencedirect.com/science/article/pii/S2468601822000074},
author = {Taitler, Ayal and Ioslovich, Ilya and Karpas, Erez and Gutman, Per-Olof},
}

@article{coles2012colin,
  title   = {COLIN: Planning with continuous linear numeric change},
  author  = {Coles, Amanda Jane and Coles, Andrew I and Fox, Maria and Long, Derek},
  journal = {Journal of Artificial Intelligence Research},
  volume  = {44},
  pages   = {1--96},
  year    = {2012}
}

@inproceedings{fernandez2015mixed,
  author       = {Enrique Fern{\'{a}}ndez{-}Gonz{\'{a}}lez and
                  Erez Karpas and
                  Brian Charles Williams},
  title        = {Mixed Discrete-Continuous Heuristic Generative Planning Based on Flow
                  Tubes},
  booktitle    = {{International Joint Conference on Artificial Intelligence}},
  pages        = {1565--1572},
  publisher    = {{AAAI} Press},
  year         = {2015}
}

@inproceedings{chen2021optimal,
  author       = {Jingkai Chen and
                  Brian C. Williams and
                  Chuchu Fan},
  title        = {Optimal mixed discrete-continuous planning for linear hybrid systems},
  booktitle    = {{International Conference on Hybrid Systems: Computation and Control.}},
  pages        = {8:1--8:12},
  publisher    = {{ACM}},
  year         = {2021}
}

@inproceedings{denenberg2019mixed,
  author       = {Elad Denenberg and
                  Amanda Jane Coles},
  title        = {Mixed Discrete Continuous Non-Linear Planning through Piecewise Linear
                  Approximation},
  booktitle    = {{International Conference on Automated Planning and Scheduling}},
  pages        = {137--145},
  publisher    = {{AAAI} Press},
  year         = {2019}
}

@article{taitler2024ipc,
  title={The 2023 International Planning Competition},
  author={Taitler, Ayal and Alford, Ron and Espasa, Joan and Behnke, Gregor and Fi{\v{s}}er, Daniel and Gimelfarb, Michael and Pommerening, Florian and Sanner, Scott and Scala, Enrico and Schreiber, Dominik and Segovia-Aguas, Javier and Seipp, Jendrik},
  journal = {AI Magazine},
  volume = {45},
  number = {2},
  pages = {280-296},
  year={2024},
  publisher={Wiley Online Library}
}

@inproceedings{taitler2017learning,
  title={Learning control for air hockey striking using deep reinforcement learning},
  author={Taitler, Ayal and Shimkin, Nahum},
  booktitle={{International Conference on Control, Artificial Intelligence, Robotics \& Optimization}},
  pages={22--27},
  year={2017},
  organization={IEEE}
}

@inproceedings{farsang2021decaying,
  author       = {M{\'{o}}nika Farsang and
                  Luca Szegletes},
  title        = {Decaying Clipping Range in Proximal Policy Optimization},
  booktitle    = {{International Symposium on Applied Computational Intelligence and Informatics}},
  pages        = {521--526},
  publisher    = {{IEEE}},
  year         = {2021}
}

@article{zhou2020graph,
  title     = {Graph neural networks: A review of methods and applications},
  author    = {Zhou, Jie and Cui, Ganqu and Hu, Shengding and Zhang, Zhengyan and Yang, Cheng and Liu, Zhiyuan and Wang, Lifeng and Li, Changcheng and Sun, Maosong},
  journal   = {AI open},
  volume    = {1},
  pages     = {57--81},
  year      = {2020},
  publisher = {Elsevier}
}

\end{document}